\documentclass[sigconf]{acmart}

\usepackage{graphicx}
\usepackage{booktabs}
\usepackage{multirow}
\usepackage{algorithm}
\usepackage{algpseudocode}
\usepackage{subcaption}

\AtBeginDocument{%
  }
\copyrightyear{2025} 
\acmYear{2025} 
\setcopyright{rightsretained} 
\acmConference[WSDM '25]{Proceedings of the Eighteenth ACM International Conference on Web Search and Data Mining}{March 10--14, 2025}{Hannover, Germany}
\acmBooktitle{Proceedings of the Eighteenth ACM International Conference on Web Search and Data Mining (WSDM '25), March 10--14, 2025, Hannover, Germany}\acmDOI{10.1145/3701551.3703541}
\acmISBN{979-8-4007-1329-3/25/03}

\begin{document}

\title{GAMED: Knowledge Adaptive Multi-Experts Decoupling for Multimodal Fake News Detection}

\author{Lingzhi Shen}
\orcid{0000-0001-6686-0976}
\affiliation{
    \institution{University of Southampton}
    \city{Southampton}
    \country{United Kingdom}
}
\email{l.shen@soton.ac.uk}

\author{Yunfei Long}
\orcid{0000-0002-4407-578X}
\affiliation{
    \institution{University of Essex }
    \city{Essex}
    \country{United Kingdom}
}
\email{yl20051@essex.ac.uk}

\author{Xiaohao Cai}
\orcid{0000-0003-0924-2834}
\affiliation{
    \institution{University of Southampton}
    \city{Southampton}
    \country{United Kingdom}
}
\email{x.cai@soton.ac.uk}

\author{Imran Razzak}
\orcid{0000-0002-3930-6600}
 \affiliation{
     \institution{Mohamed bin Zayed University of Artificial Intelligence} 
     \city{Abu Dhabi}
     \country{United Arab Emirates}
      \country{}
 }
\affiliation{
     \institution{The University of New South Wales}
     \city{Sydney}
     \country{Australia}
    \country{}
}
\email{imran.razzak@unsw.edu.au}


\author{Guanming Chen}
\orcid{0009-0002-5946-9076}
\author{ Kang Liu}
\orcid{0009-0009-6782-6126}
\affiliation{
    \institution{University of Southampton}
    \city{Southampton}
    \country{United Kingdom}
}
\email{gc3n21@soton.ac.uk}
\email{kl2y21@soton.ac.uk}


\author{Shoaib Jameel}
\orcid{0000-0001-7534-3313}
\affiliation{
    \institution{University of Southampton}
    \city{Southampton}
    \country{United Kingdom}
}
\email{M.S.Jameel@southampton.ac.uk}
\authornote{Shoaib Jameel is the corresponding author.}

\renewcommand{\shortauthors}{Lingzhi Shen et al.}

\begin{abstract}
Multimodal fake news detection often involves modelling heterogeneous data sources, such as vision and language. Existing detection methods typically rely on fusion effectiveness and cross-modal consistency to model the content, complicating understanding how each modality affects prediction accuracy. Additionally, these methods are primarily based on static feature modelling, making it difficult to adapt to the dynamic changes and relationships between different data modalities. This paper develops a significantly novel approach, GAMED, for multimodal modelling, which focuses on generating distinctive and discriminative features through modal decoupling to enhance cross-modal synergies, thereby optimizing overall performance in the detection process. GAMED leverages multiple parallel expert networks to refine features and pre-embed semantic knowledge to improve the experts' ability in information selection and viewpoint sharing. Subsequently, the feature distribution of each modality is adaptively adjusted based on the respective experts' opinions. GAMED also introduces a novel classification technique to dynamically manage contributions from different modalities, while improving the explainability of decisions. Experimental results on the Fakeddit and Yang datasets demonstrate that GAMED performs better than recently developed state-of-the-art models. The source code can be accessed at https://github.com/slz0925/GAMED.
\end{abstract}



\maketitle
\section{Introduction}
Imagine encountering a social media post with a seemingly innocuous image and a captivating new headline. Unfortunately, a sinister truth lurks beneath this alluring facade -- it is well-crafted fake news \cite{hu2024bad, gao2024knowledge}. Nowadays everyone is an editor and everyone can publish news -- especially on social media \cite{schulz2024role}. As a result, there is an escalating threat of multimodal fake news \cite{jing2023multimodal, wu2021multimodal}, a potent weapon that weaponizes the synergy of text and visuals to manipulate public discourse and erode trust in information \cite{wang2018eann, verma2023mcred}. According to \cite{zimdars2020fake}, fake news is defined as ``purposefully crafted, sensational, emotionally charged, misleading or fabricated information that mimics the form of mainstream news''.

Traditional multimodal detection \cite{amri2021exmulf}, typically rely on basic fusion techniques \cite{yang2024mran, luvembe2024caf}, striving to decipher the complex interactions within multimodal narratives \cite{lu2024fact}. The result is that they fail to capture the nuances that distinguish genuine news from its fabricated counterparts \cite{comito2023multimodal, tufchi2023comprehensive}. This critical gap in detection capabilities stems from several inherent limitations, for instance, many current fusion techniques suffer from feature suppression issues \cite{zhang2025learning}, hindering the model's ability to grasp the intricate dance between textual and visual elements within a single content. Another reason is the over-reliance on the inherent capabilities of pre-trained models without any additional refinement step for feature representations in the pipeline \cite{alghamdi2023towards, praseed2023hindi}, which is prone to lose information that is crucial for classification. Similarly, existing methods often focus solely on identifying consistency between modalities \cite{sun2023inconsistent}, and they fail to consider utilizing discriminative features as a complement. However, in the real world, many fabricators have learned to bypass the detection of previous consistency-centric models \cite{Aleroud2020BypassingDO, Meel2021HANIC}. Moreover, the black-box nature of decision-making processes in existing models shrouds the perspectives and contributions of each modality in mystery. This lack of transparency hinders interpretability \cite{guo2023interpretable, shu2019defend} and erodes user trust \cite{qiao2020language}.

Our primary objective is to create a novel approach called GAMED, which autonomously models fake news within a multimodal context, enhancing the current state-of-the-art by overcoming some of its intrinsic limitations. In pursuit of this goal, we utilize publicly accessible multimodal datasets. For example, \cite{ying2023bootstrapping} presented the BMR model for multimodal fake news detection. While BMR promotes in-depth multimodal analysis and improved feature extraction, it suffers from some key shortcomings. Firstly, BMR faces challenges in integrating multimodal data and dynamically weighting different modalities. Additionally, the bootstrapping process for multi-view representations hinders BMR's interpretability. Most importantly, BMR fails to invoke valuable real-world factual knowledge as a reference when the model encounters confusion \cite{hu2021compare, ma2023kapalm}. Another recent work by \citet{xuan2024lemma} investigated multimodal fake news detection with their LEMMA system which combines vision language models with external knowledge, but it lacks the feature refinement strategy, and this may also introduce additional noise.

\textbf{Technical contributions:} GAMED, based on modal-decoupling modelling, exploits the potential of cross-modal synergies to improve detection performance, which is distinct from existing multimodal fake news detection methods based on consistency learning or fusion-only strategies, such as those in \cite{song2022dynamic, haghir2024review, lakzaei2024disinformation}. GAMED combines the characteristics of expert networks and AdaIN \cite{huang2017arbitrary} to perform progressive feature refinement to obtain more discriminative and distinctive feature representations, providing a novel paradigm for dynamic screening and optimization of multimodal data. GAMED demonstrates that external knowledge, e.g., semantic knowledge graph information encoded in pre-trained language models \cite{fang2022dtcrskg} is beneficial to help models better understand the complex relationships and contexts in fake news, and extends with the influence from text to other modalities. GAMED introduces a novel decision-making method, which is conducive to improving the transparency and explainability of the decision-making process. GAMED attains better detection performance on the publicly available Fakeddit\footnote{https://github.com/entitize/Fakeddit} and Yang datasets \cite{yang2018ti}, presenting a novel solution for automated fake news detection.

\begin{figure*}
    \centering
    \includegraphics[scale=0.220]{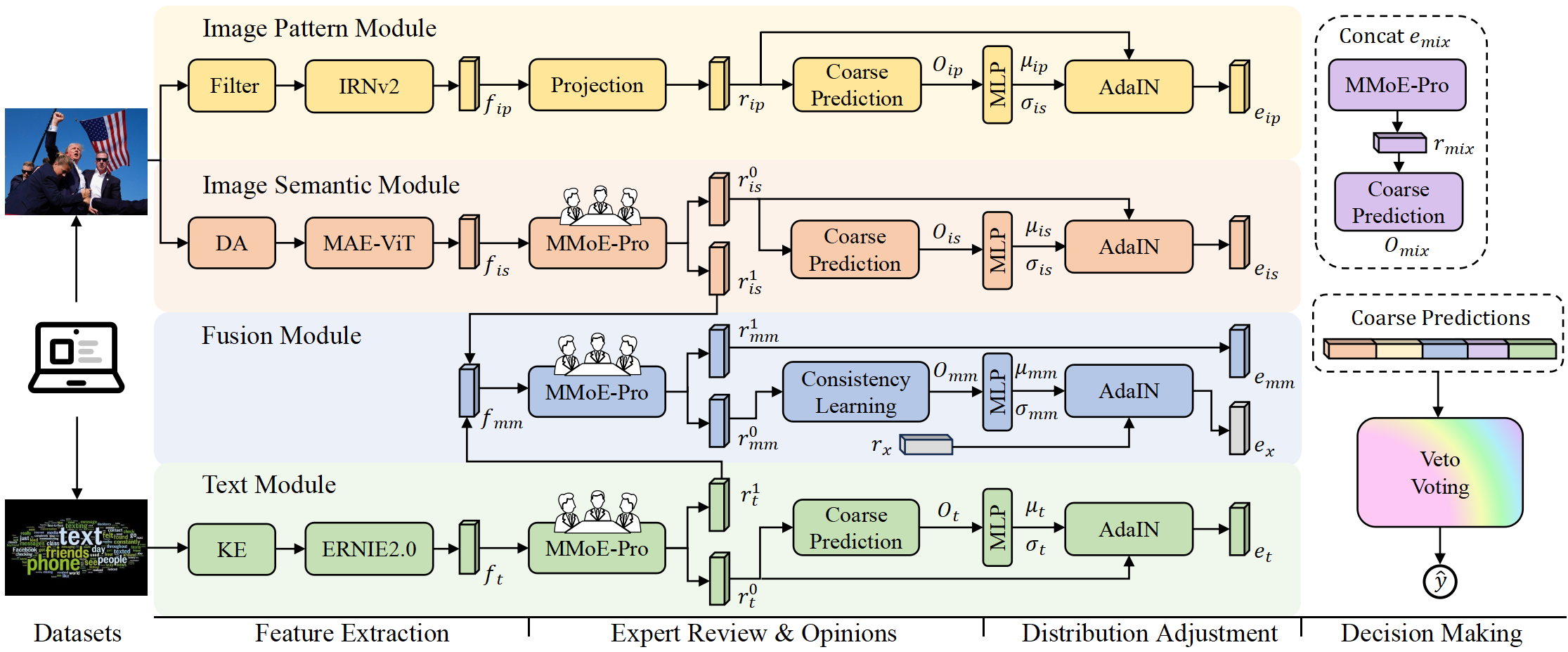} %
    \caption {Starting from raw data, the GAMED's modality-specific pipeline performs feature extraction and progressive refinement. The knowledge enhancement mechanism provides an external background to the architecture. During the expert review stage, features are selected and coarse predictions are made. The AdaIN component then adaptively adjusts the feature distribution. The decisive voting stage orchestrates the final classification.}
    \label{fig:model-architecture}
\end{figure*}

\section{Related Work}
\noindent \textbf{Unimodal fake news methods:} Unimodal fake news detection \cite{chen2022cross, gangireddy2020unsupervised, shim2021link2vec} has made significant strides in tackling fake news \cite{silva2021embracing}. Traditional machine learning algorithms, such as SVM, Decision Trees, and Na\"ive Bayes \cite{Asha2024AnEA, Jaiswal2024ENHANCINGGI, Hakak2021AnEM}, along with modern deep learning approaches, including CNN, RNN, and LSTM \cite{Hu2022DeepLF, Li2022ABS}, have been extensively compared and studied. For text analysis, transformer-based models such as BERT \cite{Devlin2019BERTPO} have paved the way for advancements such as RoBERTa and GPT-3 \cite{anirudh2023multilingual}, enhancing the detection of subtle linguistic cues. Similarly, sophisticated architectures such as ResNet and ViTs \cite{jiang2024application} have revolutionized image analysis, enabling the identification of manipulated visuals. However, unimodal approaches have an inherent weakness: they struggle against multimodal fake news that blends text, images, and other media for a more convincing narrative \cite{zhou2023multimodal}.

\noindent \textbf{Multimodal fake news methods:} Multimodal fake news detection has emerged as a critical research area, which analyzes data from multiple sources -- text, images, videos, and social context -- to form a more holistic view of the information \cite{hua2023multimodal}. By leveraging the complementary strengths, these approaches aim to uncover discrepancies that might not be evident when analyzing text alone. This field has gravitated towards leveraging the synergistic potential of advanced fusion techniques and models such as VisualBERT \cite{kumari2023identifying}, ViLBERT \cite{gautam2023fake}, and LXMERT \cite{tan2019lxmert}, which facilitate dynamic, context-aware integration of text and images through self-attention mechanisms. These models have significantly advanced the capacity to understand and analyze the complex interplay between modalities, often focusing on exploiting cross-modal dynamics and consistencies as potent indicators of misinformation. Moreover, the integration of external knowledge sources \cite{fu2023kg, liu2024entity}, through methods like knowledge graph embeddings \cite{gao2024knowledge}, has provided additional context for verifying claims, enhancing the models' ability to discern truth from deception. Despite these advancements, multi-modal detection still faces notable challenges, particularly in processing effectiveness and the adaptive generalization to new forms of fake content, such as deepfakes. The quest for explainability \cite{zhao2024explainability} in these complex models remains an ongoing challenge. Moreover, existing multimodal research predominantly focuses on innovative fusion techniques, while how to leverage the distinctive potential of each modality remains an unresolved issue \cite{lin2022modeling, lu2024coordinated}.

\section{Our Novel GAMED Model}\label{our-model}
As depicted in Figure~\ref{fig:model-architecture}, GAMED is a novel modality-decoupling design for detecting fake news across textual and visual modalities. The process starts with extracting features from text and images, followed by a stage that simulates expert review and opinions using the MMoE-Pro network to refine feature representations. Subsequently, the distribution adjustment stage, guided by the AdaIN adaptive mechanism, dynamically fine-tunes the impact of each modality, giving precedence to the most pertinent and trustworthy data. Finally, a novel voting system with veto power is introduced in the decision-making stage by combining consensus-based and confidence-based evaluation methods. The entire GAMED workflow also benefits from the semantic information encoded (KE) by pre-trained language models that encode structured (e.g., knowledge graphs) and unstructured text information. In Algorithm~\ref{alg:GAMED}, we meanwhile present the detailed pseudo-code of GAMED.

\begin{figure*}
    \centering
    \includegraphics[scale=0.212]{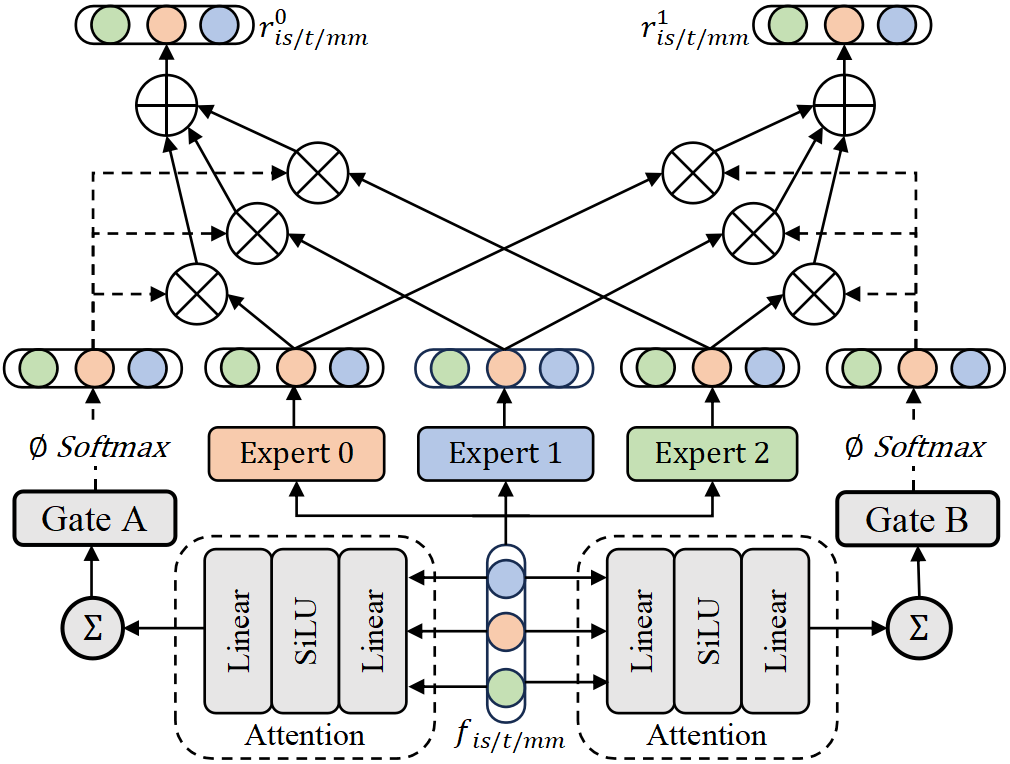}%
    \hspace{0.5cm}
    \includegraphics[scale=0.220]{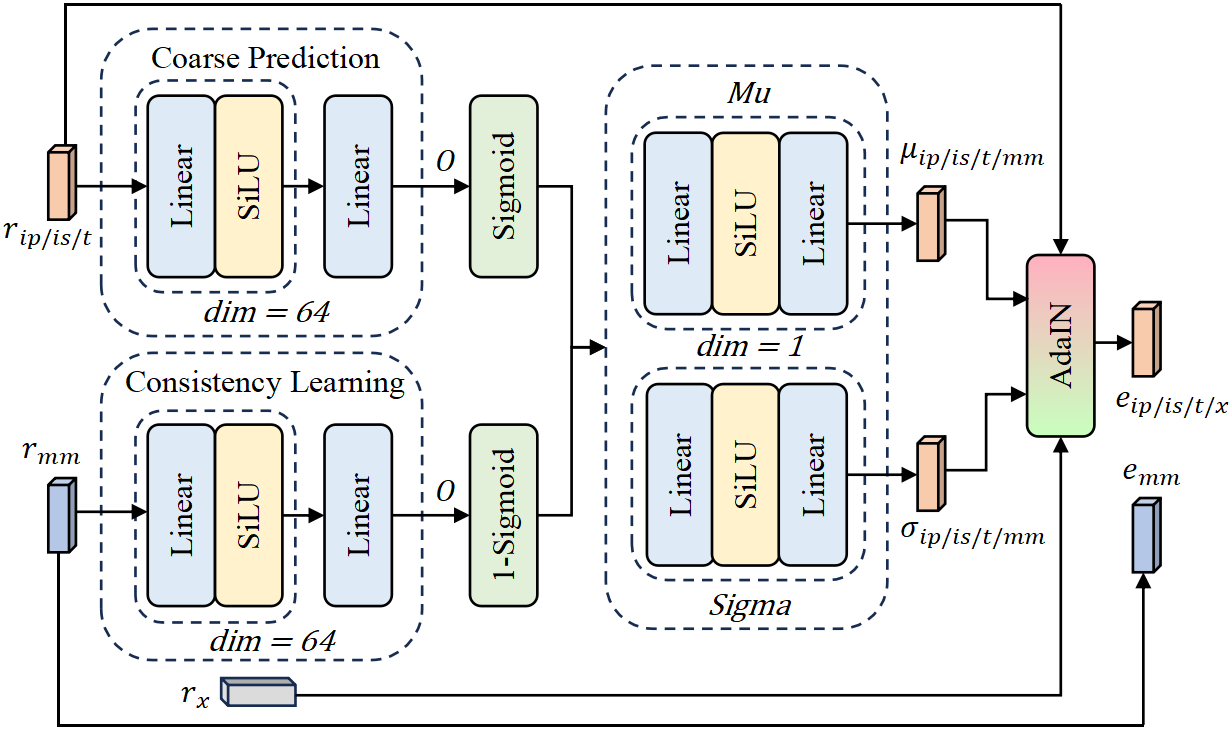}
    \caption{\textbf{Left:} The configuration of MMoE-Pro and the flow of processing representations. \textbf{Right:} The pipeline of four modules from coarse prediction to adaptive feature distribution adjustment to obtain enhanced representations.}
    \label{fig:disentangling-representation}
\end{figure*}

\noindent \textbf{Feature Extraction:} We represent our multimodal news data as a collection \(\mathcal{N}=[\mathbf{I,\mathbf{T}}] \in \mathcal{D}\), where \(\mathbf{I}, \mathbf{T}, \mathcal{D}\) are the image, the text, and the dataset, respectively. Each data point within \(\mathcal{D}\) allows us to analyze the interplay between visual content and textual narrative. We exploit the Inception-ResNet-v2 (IRNv2) \cite{szegedy2017inception} as a feature extractor to extract image patterns (IP), denoted by \(f_{ip}\). We add a special filter BayarConv \cite{bayar2018constrained} as an early layer. Our intuition is that when images are tampered with, they often leave subtle traces of forgery that are not easily detected by traditional convolutions, such as artefacts, lighting, and texture. We capture image semantic (IS) features on the global and local details of the image by combining ViT with a masked auto-encoder model (MAE) \cite{li2022semmae, gupta2024siamese}, denoted as \(f_{is}\). The use of data augmentation (DA), such as rotation, flipping, and scaling, enables the model to better generalize inconsistencies often encountered in fake images to enhance the robustness and diversity of image semantic data.

We exploit the ERNIE2.0 \cite{sun2020ernie} model to extract the text (T) representations, denoted as \(f_{t}\). ERNIE2.0 includes several advantages such as modelling sentence-level relations (in addition to word-level), and large-scale semantic knowledge stored in knowledge graphs. With these improvements, ERNIE2.0 can better evaluate the consistency of text content with visual information in images, and analyze facts and entity relationships in text. The structured knowledge encoded in ERNIE2.0 helps enhance the reasoning ability of the entire architecture in global synergies across modalities.

\noindent \textbf{Expert Review and Opinions:} The workflow of the novel expert network is depicted in the model in Figure~\ref{fig:disentangling-representation} (left). In this stage, we simulate the scenario of experts with rich expertise to review. The expert networks of different modalities accept features provided by the extractors of their respective modules, and then the mixture of experts will jointly review and select these features, and provide preliminary predictions. 

Our MMoE-Pro upgrades the traditional MMoE \cite{ma2018modeling} mainly by introducing token attention and relaxing the softmax constraint in its gating mechanism. In particular, suppose the input $f$, which can be from IS, T, or fusion (MM) modules, consists of multiple tokens, the process begins by calculating the importance score $\alpha_i$ for each token representation using a shared-weight MLP $A$, formalized as $\alpha_i = A(\text{token}_i)$. The normalized importance score $\beta_i$ is calculated as \(\beta_i = \frac{\alpha_i}{\sum_{j} \alpha_j}\). This normalization allows for the aggregation of token representations into a unified form, where the aggregated representation $\tilde{f}$ is computed as a weighted sum of the token representations: $\tilde{f} = \sum_{i} \beta_i \cdot \text{token}_i$. This approach enhances the model's ability to identify key features by dynamically evaluating the importance of different input features, thereby optimizing the feature-sharing process in multi-task learning. 

Furthermore, we adjust the gating mechanism by lifting the strict positivity and normalization constraints traditionally imposed by softmax. Specifically, the new weights $w_{t,i}(f)$, used for determining the contribution of each expert $i$ for task $t$, are derived directly from the raw scores, allowing weights to take on negative values or exceed one. It is because there is no necessity to use constraints to force the expert's contribution to be positive in our setting. The final output of the MMoE-Pro model is now formalized as
\begin{equation}
\text{MMoE-Pro}_t(f) = \sum_{i=1}^N w_{t,i}(\tilde{f}) E_i(f).
\end{equation}
Here, $E_i(f)$ represents the output of the $i$-th expert. Ultimately, the output features $\mathit{r} = \text{MMoE-Pro}_t(f)$ serve as the feature representation jointly selected by the expert team, including $\left[r_{is}^0, r_{is}^1\right]$, $\left[r_t^0, r_t^1\right]$ and $\left[r_{mm}^0, r_{mm}^1\right]$. Here, $r_{is}^1$ and $r_t^1$ serve as the initial features for the fusion module, $f_{\text{mm}}$, and as inputs to the expert network in this module. These new improvements can more flexibly and effectively enhance the ability of the expert network to dynamically allocate computing resources across different tasks and modalities.

We use an MLP classifier to accept the refined representations from the MMoE-Pro and Projection (\(f_{ip}\)) to achieve the coarse prediction function. This is depicted in the steps of coarse prediction and consistency learning in Figure~\ref{fig:disentangling-representation}. For each refined representation $\mathit{r}$ from the expert network, we perform a 64-dimensional feature reduction and produce the classification output. This combined process can be expressed as \(\mathit{O} = \text{MLP}(\mathit{r})\), where \(\mathit{O}\) represents the coarse prediction output.

\noindent \textbf{Distribution Adjustment:} We depict this stage in Figure~\ref{fig:disentangling-representation} (right). The coarse prediction results from the previous stage are calculated as mean and standard deviation as an acceptable input form for AdaIN. AdaIN then adaptively adjusts the feature distribution according to the contribution of each modality. This step ensures that the most relevant and reliable information is prioritized.

We first calculate the parameters, mean \(\mu\) and standard deviation \(\sigma\), required by AdaIN. Unlike the standard AdaIN approach, our \(\mu\) and \(\sigma\) are generated through MLP networks rather than being directly extracted from the style features. Specifically, for each output \(O\) from the coarse prediction, we use MLPs to generate the mean and standard deviation, denoted as \(\mu = \text{MLP}_{\mu}(\text{sigmoid}(O))\) and \(\sigma = \text{MLP}_{\sigma}(\text{sigmoid}(O))\), where \(\text{MLP}_{\mu}\) and \(\text{MLP}_{\sigma}\) represent the MLPs for calculating the mean and standard deviation, respectively. The adjustment process of AdaIN is then formalized as
\begin{equation}
e = \sigma ({r - \mu_r})/{\sigma_r} + \mu.
\end{equation}
Here, \(\mu_r\) and \(\sigma_r\) are respectively the mean and standard deviation of the input feature \(r\), and $\mathit{e}$ is the adjusted feature. The process of combining information from each modality in AdaIN can be simplified as \(\mathit{e} = \text{AdaIN}(\mathit{r}, \mu, \sigma)\). At this stage, for the mean and standard deviation calculation of the prediction output from consistency learning, we use (\(1 - \text{sigmoid}(\mathit{O})\)) to invert it so that adjust the distribution of irrelevant representation \( \mathit{r}_{\mathit{x}} \) to \( \mathit{e}_{\mathit{x}} \). Our intuition is that certain irrelevant information, such as emotional language, lengthy background introduction, complex rhetoric, etc., can also serve as clues for detecting fake news. Therefore, we decouple relevance from consistency learning and leverage irrelevance as a synergistic supplement. We directly use the fusion representation \( \mathit{r}_{\mathit{mm}} \) as the adjusted representation \( \mathit{e}_{\mathit{mm}} \) of the fusion module.

\begin{figure}
    \centering
    \includegraphics[scale=0.178]{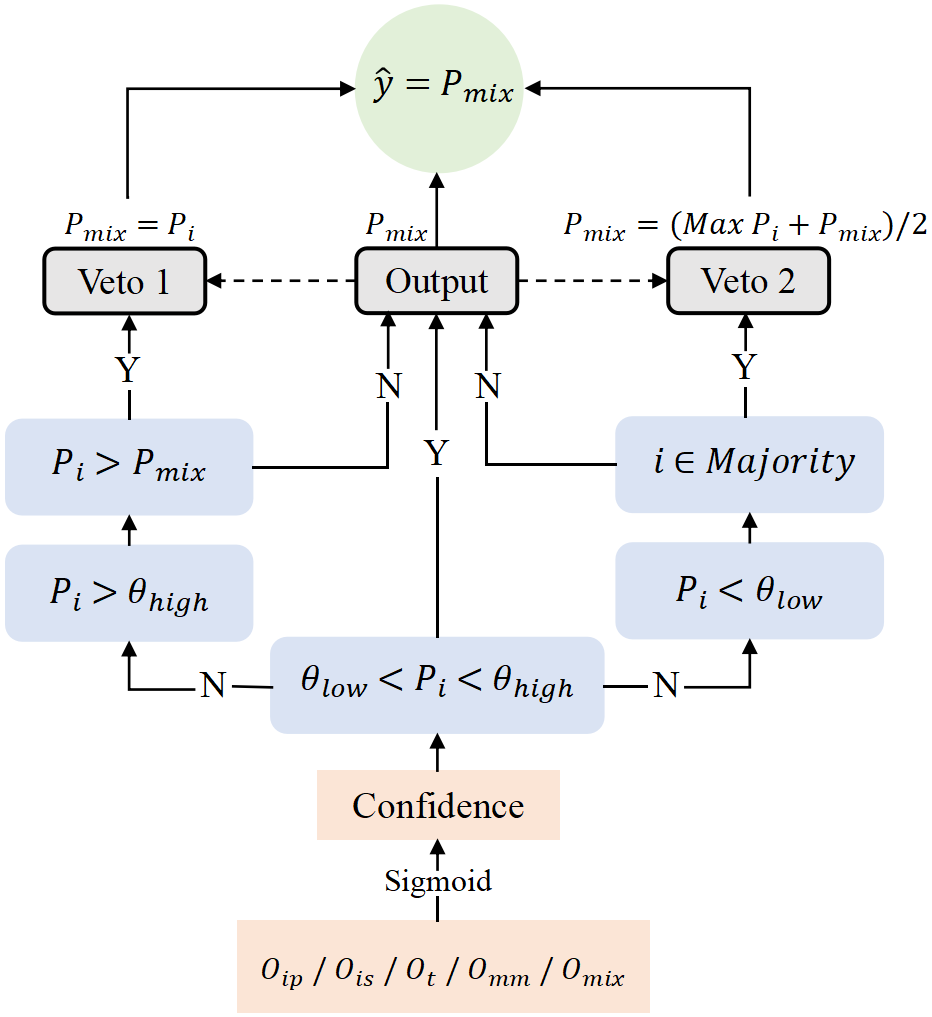}
    \caption{Our novel veto model.}
    \label{fig:veto-voting}
\vspace{-0.10in}
\end{figure}

\noindent \textbf{Veto Voting:} The novel veto classifier is depicted in Figure~\ref{fig:veto-voting}. Before the voting mechanism is triggered, the enhanced representations of all modalities produced by AdaIN are concatenated and denoted as \(\mathit{e_{mix}}\) (Figure~\ref{fig:model-architecture}), and then the MMoE-Pro performs the same refinement process to obtain \(\mathit{r_{mix}}\) (Figure~\ref{fig:model-architecture}) and \( \mathit{O_{mix}} \) successively. Finally, this concatenated prediction output and the previous coarse prediction output of each modality are used as input into the voting stage. Our novel veto voting combines the inspiration of thresholds and confidence to dynamically manage the contributions and conflicts of each modality prediction to ensure the reliability and transparency of the decision-making.

We define two thresholds to distinguish between high confidence and low confidence, where \( \theta_{\text{high}} \) and \( \theta_{\text{low}} \) define the high confidence and low confidence thresholds, respectively, used to determine if a module's prediction can be used as a decision basis. For the prediction output \(O_i\) of each module \(i\), we apply the sigmoid function to convert it into a confidence probability value as \( \mathit{P}_i = \mathrm{sigmoid}(\mathit{O}_i) \) where \(P_i\) is the confidence of the prediction from module \(i\). Let the confidence of the concatenated output be \( P_{\mathit{mix}} \), initially set as \( \mathit{P}_{\mathit{mix}} = \mathrm{sigmoid}(\mathit{O}_{\mathit{mix}}) \). Suppose \( P_{\mathit{mix}} \) is used as the basis for comparing other module confidences and for the final decision. Let \(O_i\) and \( O_{\mathit{mix}} \) be the raw outputs of the module prediction and the concatenated output, respectively. Let the majority class be the class decided by the majority of module decisions; whether module \(i\) belongs to the majority class can be determined by comparing it to other module predictions. We iterate through each module and update the concatenated output according to the following rules.

\noindent \textbf{Rule 1}: For the initial concatenated output, we denote the rule as \( \mathit{P}_{\mathit{mix}} = \text{sigmoid}(\mathit{O}_{\mathit{mix}}) \).

\noindent \textbf{Rule 2}: If the module confidence \( P_i \) is greater than the high confidence threshold and \( P_i > P_{\mathit{mix}} \), then replace the concatenated output with the output from module \( i \). This is denoted as \( P_{\mathit{mix}} = P_i \ \text{if} \ P_i > \theta_{\text{high}} \ \text{and} \ P_i > P_{\mathit{mix}} \).

\noindent \textbf{Rule 3}: If the module confidence \( P_i \) is less than the low confidence threshold, and module \(i\) belongs to the majority class, then ignore any output in the majority class, and reconsider the maximum output from all modules. This is denoted as \( P_{\mathit{mix}} = \frac{1}{2} \left(P_{\mathit{mix}} + \max P_i\right) \ \text{if} \ P_i < \theta_{\text{low}} \ \text{and } i \ \text{belongs to the majority class} \).

\noindent \textbf{Rule 4}: If the confidence is between the thresholds, maintain the concatenated output, denoted as \( P_{\mathit{mix}} = P_{\mathit{mix}} \ \text{if} \ \theta_{\text{low}} \leq P_i \leq \theta_{\text{high}} \).

\begin{algorithm}[t]
\caption{PyTorch-style Pseudocode of GAMED}
\label{alg:GAMED}
\begin{algorithmic}[1]
\Require Dataset: $\mathcal{D}$; training epochs: $N$; batch size: $B$; and learning rate: $\eta$.
\Ensure Model parameters: $\Theta$
\State Define loss function: $\mathcal{L}_{\text{BCE}}$
\State Define optimizer: AdamW
\For {epoch in range($N$)}
    \State Load batch data $(T, I, y)$
    \State \textbf{Forward Propagation:}
    \State \hspace{1em} $f = t, is, ip = \text{extract}(T, I)$
    \State \hspace{1em} $r = \text{experts}(t, is, ip)$
    \State \hspace{1em} $r_{mm} = \text{experts}(mm = t + is)$
    \State \hspace{1em} $O = \text{class}(\text{reduc}(r_0, r_1, \ldots, r_n))$
    \State \hspace{1em} $\mu, \sigma = \text{comp}(O, 1 - O)$
    \State \hspace{1em} $e = \text{AdaIN}(r, \mu, \sigma)$
    \State \hspace{1em} $f_{mix} = \text{concat}(e_0, e_1, \ldots, e_n)$
    \State \hspace{1em} $r_{mix} = \text{experts}(f_{mix})$
    \State \hspace{1em} $O_{mix} = \text{class}(\text{reduc}(r_{mix}))$
    \State \hspace{1em} $\hat{y} = \text{veto}(O_0, O_1, \ldots, O_n)$
    \State \textbf{Back Propagation:}
    \State \hspace{1em} $\mathcal{L} = \text{ComputeLoss}(O_0, O_1, \ldots, O_n)$
    \State \hspace{1em} $\mathcal{L}.\text{backward}()$
    \State \hspace{1em} optimizer.step()
\EndFor
\end{algorithmic}
\end{algorithm}

\section{Experiments and Results}
To rigorously evaluate the efficacy of GAMED in detecting fake news, we conducted extensive experiments. This section details the experimental framework, evaluation criteria, and the notable results obtained. The objective is to determine whether GAMED outperforms recent robust models and to assess the contribution of each component through ablation studies. A qualitative analysis also demonstrates the transparency of the decision-making process.

\subsection{Experimental Setup}
\noindent \textbf{Datasets:} We conducted training, validation, and test of GAMED and other models on two publicly accessible datasets: Fakeddit and Yang. Fakeddit is a vast collection with over one million labelled samples, classified as real or fake news. It offers a balanced division, consisting of 628,501 fake news instances and 527,049 real news instances. Derived from a wide range of 22 subreddits, Fakeddit provides a rich diversity of domains and topics, mirroring the real-world scenario. Fakeddit is a dataset that offers fine-grained categories. For our model, which focuses on binary classification tasks, we utilize only the 2-way labels.
The Yang dataset includes 20,015 news articles, with 11,941 marked as fake and 8,074 as real. The fake news is sourced from more than 240 websites, while the genuine news is obtained from reputable, authoritative outlets like the New York Times and Washington Post. The dataset used in \citet{ying2023bootstrapping} is not publicly accessible due to strict API restrictions on obtaining image data from Twitter and Weibo. Our attempts to contact the authors were unsuccessful.

\noindent \textbf{Settings:} The Fakeddit dataset comprises 563,612 training samples, 58,798 validation samples, and 59,271 test samples. In contrast, Yang's dataset contains 4,655 training samples, 582 validation samples, and 583 test samples. Each text has a corresponding image. In processing image data, we utilize two pre-trained models: \textit{mae-pretrain-vit-base} for semantic analysis and \textit{pytorch-InceptionResNetV2} for pattern recognition. Text data is processed with \textit{ernie-2.0-base-en}. MAE-ViT and ERNIE have a hidden dimension of 768, with their parameters kept frozen. Our preprocessing steps aim to optimize the handling of input data. This process includes resizing all images to a consistent size of $224\times224$ pixels. We also set a maximum tokenization length of 197 for both text and image data. All MLPs in GAMED include one hidden layer and SiLU activation function. We use AdamW optimizer with $1\times10^{-4}$ learning rate. The model typically reaches peak accuracy within 9--10 epochs on Fakeddit and 5--7 epochs on Yang.

\noindent \textbf{Evaluation Metrics:} To align with comparative models, we use the widely accepted metrics: Accuracy (Acc), Recall (R), Precision (P), and F1 Score for this task.

\noindent \textbf{Comparative Models:} The selection of baselines is shown in Table~\ref{tab:state-of-the-art}. For the Fakeddit dataset, our model is benchmarked against recent robust comparative models. The EANN model \cite{wang2018eann} combines Text-CNN and VGG-19 to filter event-specific features while retaining event-independent features, enhancing adaptability to new events. The MVAE \cite{khattar2019mvae} model was chosen for its use of a multimodal variational autoencoder that learns joint representations from text and image data, improving detection by capturing the interactions between modalities with LSTM and VGG-19. The BMR model \cite{ying2023bootstrapping}, on the other hand, bootstraps multi-view representations to refine and reweigh features from text and image for superior fake news detection performance. The MMBT \cite{kiela2019supervised} integrates text features from BERT with image features from ResNet-152 using a single transformer, while MTTV \cite{wang2023multi} employs a dual-level visual feature extraction approach with BERT and ResNet to bolster the synergy between textual and visual data. The CLIP and LLaVA hybrid architecture \cite{Lee2024HowTT} employs LoRA-based fine-tuning strategies and knowledge transfer, effectively enhancing multimodal fact verification by integrating visual and textual evidence. ELD-FN \cite{luqman2024utilizing} combines ViBERT-generated multimodal embeddings with sentiment analysis through an ensemble learning approach.

For the Yang dataset, in addition to using the same baselines (i.e., EANN, MVAE, and BMR) as for the Fakeddit dataset, we further introduced several advanced architectures. For instance, the MCNN \cite{xue2021detecting} model integrates text and visual features through five sub-networks, including BERT, BiGRU, ResNet50, ELA algorithms, and attention mechanisms, detecting inconsistencies in multimodal data by measuring similarity and identifying visual tampering. The SAFE \cite{zhou2020similarity} model extracts features from news text and visuals using an extended Text-CNN and Image2Sentence model, identifying mismatches between text and images. TI-CNN \cite{yang2018ti} uses parallel CNNs to extract and fuse explicit and latent features from text and images, establishing itself as an initial benchmark for the Yang dataset. Finally, MVNN \cite{qi2019exploiting} combines frequency and pixel domain visual information using CNN and multi-branch CNN-RNN networks, along with attention mechanisms to dynamically integrate physical and semantic features in fake news images.

\begin{table}
    \centering
    \scalebox{0.880}
    {
    \begin{tabular}{cccccc}
        \toprule
        Dataset & Method & Acc & P & R & F1 \\ 
        \midrule
        \multirow{8}{*}{Fakeddit}
        & EANN & 87.50 & 90.43 & 88.11 & 89.26 \\
        & MVAE & 88.75 & 90.11 & 91.39 & 90.74 \\
        & ELD-FN & 88.83 & 93.54 & 90.29 & 91.89 \\
        & MMBT & 91.11 & 92.74 & 92.51 & 92.63 \\
        & MTTV & 91.88 & 93.48 & 93.03 & 93.25 \\
        & BMR & 91.65 & \textbf{94.34} & 92.88 & 93.61 \\
        & CLIP+LLaVA & 92.54 & 93.85 & 91.24 & 92.53 \\
        & \textbf{GAMED} & \textbf{93.93} & 93.55 & \textbf{93.71} & \textbf{93.63} \\ 
        \midrule
        \multirow{8}{*}{Yang} 
        & EANN & 85.54 & 86.37 & 84.32 & 85.33 \\
        & MVAE & 90.85 & 91.58 & 90.94 & 91.26 \\
        & SAFE & 92.27 & 93.75 & 93.53 & 93.64 \\
        & TI-CNN & 92.48 & 92.20 & 92.77 & 92.10 \\
        & BMR & 94.34 & 95.23 & 93.59 & 94.15 \\
        & BERT+MVNN & 95.68 & 96.44 & 95.93 & 96.18 \\
        & MCNN & 96.30 & 97.29 & 96.44 & 96.86 \\
        & \textbf{GAMED} & \textbf{98.46} & \textbf{98.31} & \textbf{98.59} & \textbf{98.43} \\
        \bottomrule
    \end{tabular}
    }
    \caption{Comparison between our GAMED and state-of-the-art. The best results (\%) are highlighted in bold.}
    \label{tab:state-of-the-art}
    \vspace{-0.20in}
\end{table}

\begin{table}
    \centering
    \scalebox{0.868}
    {
    \setlength{\tabcolsep}{4pt}
    \begin{tabular}{lccccccc}
        \hline
        \multicolumn{8}{c}{Fakeddit Dataset} \\ \hline
        \multirow{2}{*}{Method} & \multirow{2}{*}{Acc} & \multicolumn{3}{c}{Fake News} & \multicolumn{3}{c}{Real News} \\ \cline{3-8} 
        &  & P & R & F1 & P & R & F1 \\ \hline
        LLaVA (Direct) & 0.663 & 0.588 & 0.797 & 0.677 & 0.777 & 0.558 & 0.649 \\
        LLaVA (CoT) & 0.673 & 0.612 & 0.400 & 0.484 & 0.694 & 0.843 & 0.761 \\
        GPT-4 (Direct) & 0.677 & 0.598 & 0.771 & 0.674 & 0.776 & 0.606 & 0.680 \\
        GPT-4 (CoT) & 0.691 & 0.662 & 0.573 & 0.614 & 0.708 & 0.779 & 0.742 \\
        GPT-4V (Direct) & 0.734 & 0.673 & 0.723 & 0.697 & 0.771 & 0.742 & 0.764 \\
        GPT-4V (CoT) & 0.754 & 0.858 & 0.513 & 0.642 & 0.720 & \textbf{0.937} & 0.814 \\
        FacTool & 0.506 & 0.476 & 0.834 & 0.606 & 0.624 & 0.232 & 0.339 \\
        InstructBLIP & 0.726 & 0.760 & 0.489 & 0.595 & 0.715 & 0.892 & 0.793 \\
        LEMMA & 0.824 & 0.835 & 0.727 & 0.777 & 0.818 & 0.895 & 0.854 \\
        \hline
        \textbf{GAMED} & \textbf{0.939} & \textbf{0.954} & \textbf{0.944} & \textbf{0.949} & \textbf{0.917} & 0.930 & \textbf{0.923} \\ \hline
    \end{tabular}
    }
    \caption{Comparison between GAMED and large language models on Fakeddit. The best results are highlighted in bold.}
    \label{tab:language-models}
    \vspace{-0.15in}
\end{table}

\subsection{Overall Results}
The results in Table~\ref{tab:state-of-the-art} demonstrate that our GAMED is quantitatively superior in performance when compared with different competitive models including those that are recently developed. On Fakeddit, GAMED achieves 93.93\% accuracy, surpassing the state-of-the-art open-source detection scheme MTTV by 2.05\% and LoRA-fine-tuned CLIP and LLaVA combination by 1.39\%. On the Yang dataset, GAMED achieved a remarkable accuracy of 98.46\%, surpassing the state-of-the-art MCNN by 2.16\%. Meanwhile, GAMED also ranks first in Precision, Recall, and F1 on both Fakeddit and Yang. We tested recently developed BMR architecture on both Fakeddit and Yang datasets, and the results demonstrate that our GAMED outperforms BMR in almost all evaluation metrics, except for the Precision of Fakeddit.

Given the popularity of large language models (LLMs), in Table ~\ref{tab:language-models}, we compare GAMED with LLMs-based fake news detection schemes, and the experiments are conducted on Fakeddit and depicted in Table~\ref{tab:language-models}. We obtained similar conclusions on the Yang dataset too. The Direct approach employs the model for fake news detection without any preprocessing of the input data, relying solely on the model's internal knowledge to directly generate predictions and reasoning. In contrast, the Chain of Thought (CoT) \cite{wei2022chain} approach enhances the model's ability to handle complex tasks by prompting it to ``think step by step'', guiding the model to first produce a reasoning process before delivering a final prediction. Both LLaVA \cite{li2024llava} and the GPT-4 family \cite{achiam2023gpt} underperformed on this task, falling short of most traditional language models listed in Table ~\ref{tab:state-of-the-art}. Furthermore, fine-tuned LLMs like InstructBLIP \cite{Dai2023InstructBLIPTG} or tool-enhanced LLMs like FacTool \cite{chern2023factool} did not improve the capabilities of LLMs. The state-of-the-art architecture LEMMA \cite{xuan2024lemma} has an accuracy of only 82.4\%, i.e., significantly lower than our GAMED by 11.5\%. In Precision, Recall and F1 score, GAMED shows improvement better than LLMs, except that it ranks second in Recall for the real news category, slightly lower than GPT-4V using CoT.

Our model performs better than the strong comparative models with several reasons. As mentioned before, we address some of the key shortcomings in the existing models. By employing modal decoupling and cross-modal synergy, GAMED preserves and enhances the discriminative features of each modality. The feature refinement components dynamically emphasize the most relevant information, in contrast to the static methods used before. Moreover, GAMED exploits semantic knowledge from pre-trained models, deepening its understanding of facts and relationships, which is a capability that many comparative models lack. Finally, our novel veto voting mechanism, which combines consensus and confidence, ensures that the most reliable predictions drive the final decision, offering greater flexibility than traditional voting or fusion methods.

\begin{figure}
    \centering
    \includegraphics[scale=0.295]{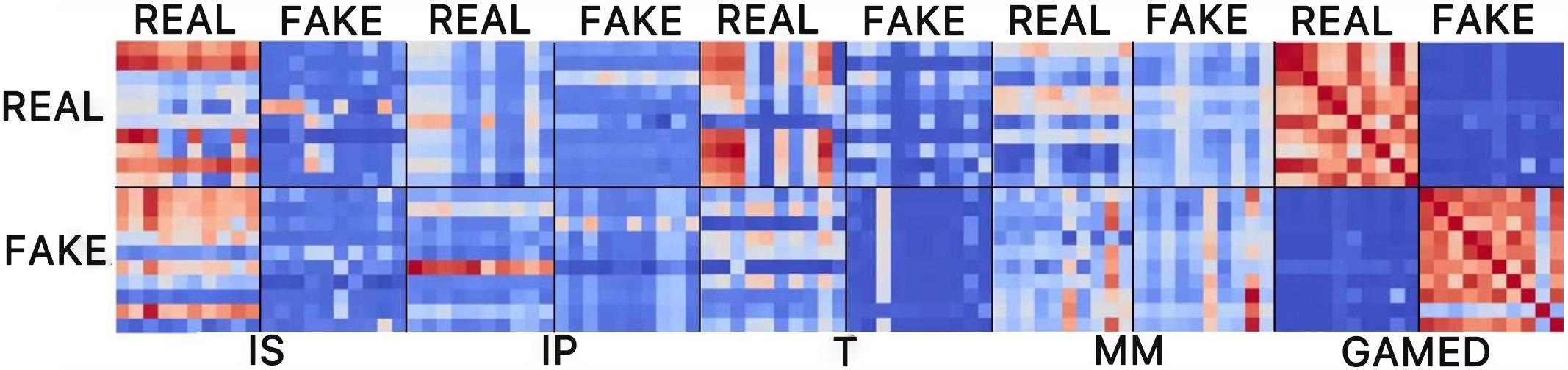} %
    \includegraphics[scale=0.314]{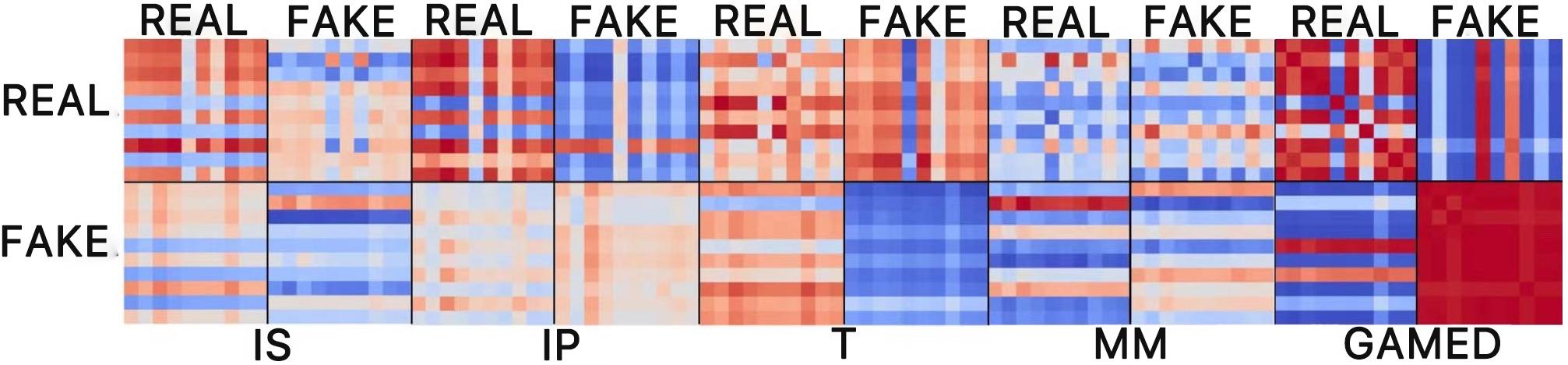}
    \caption{Heatmaps of cosine similarity on Fakeddit and Yang. Each heatmap cell shows the pairwise cosine similarity between the 64-dimensional representation from the coarse predictor of four modules and the whole model.}
    \label{fig:heat-map}
    \vspace{-4mm}
\end{figure}

As depicted in Figure~\ref{fig:heat-map}, we randomly select ten fake and ten real news samples to visualize the heat map. The colours of the heat map range from dark blue (low similarity) to dark red (high similarity) showing the degree of similarity between features. The fake news features within a cell are more scattered than the real news features, which is often the reason for the difficulty in detection. However, GAMED successfully captures these discriminative news features. The differences between modules reflect that each module can provide distinctive perspectives and information when working. This again verifies our intuition that leveraging the discriminability and distinctiveness from modal decoupling can enhance the model's overall detection performance.

\begin{figure}
    \centering
    \includegraphics[scale=0.165]{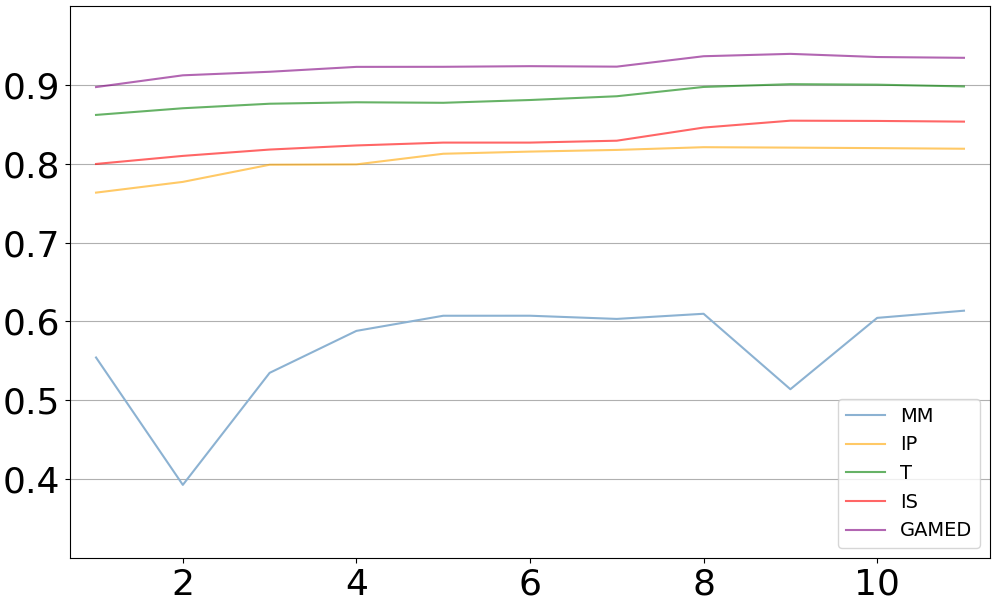} %
    \includegraphics[scale=0.165]{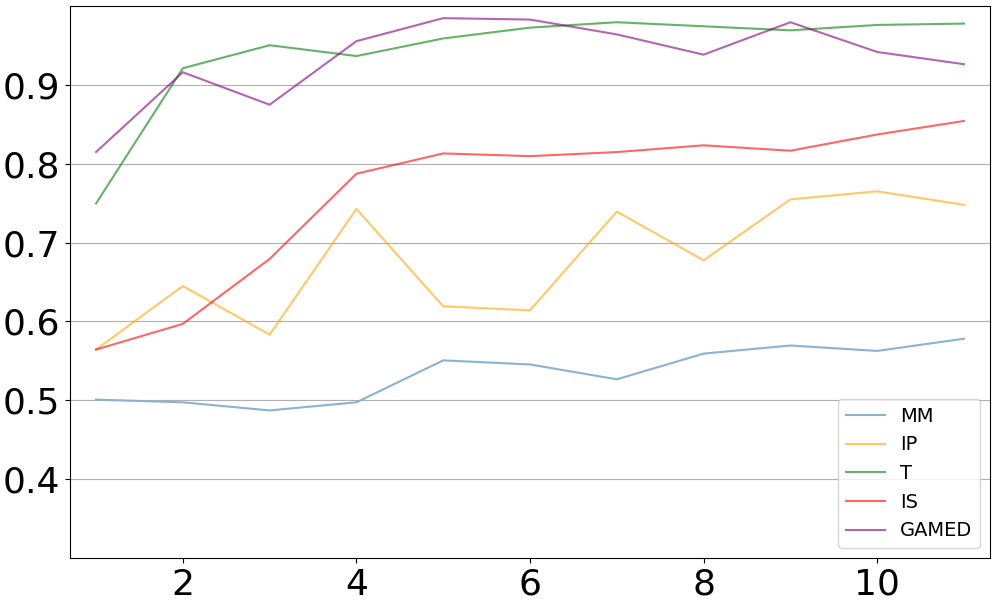} 
    \includegraphics[scale=0.145]{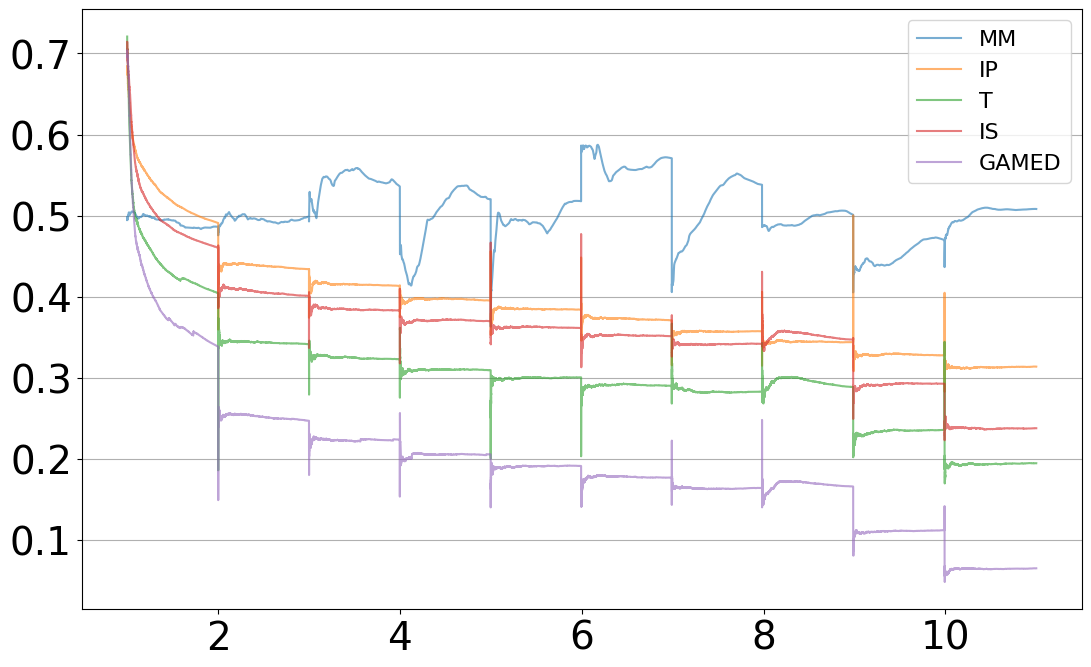} %
    \includegraphics[scale=0.145]{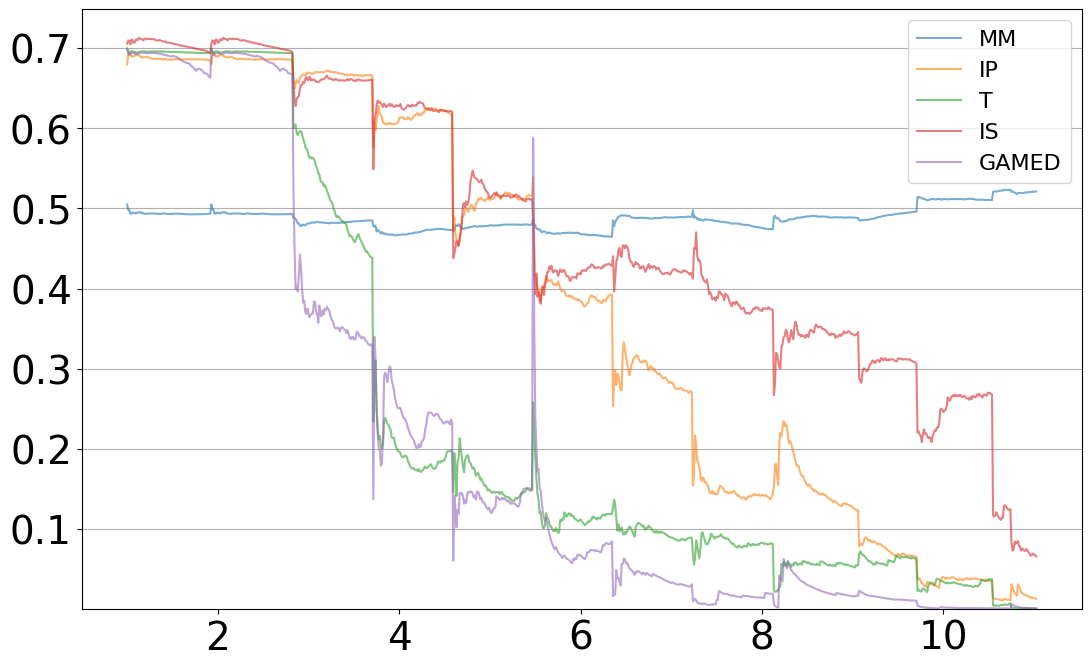}
    \caption{The comparison of accuracy (first row) and loss (second row) illustrates the learning curves of GAMED and its four modules during training. The training set on the left column and right column is Fakeddit and Yang, respectively.}
    \label{fig:loss-curve}
    \vspace{-4mm}
\end{figure}

Figure \ref{fig:loss-curve} illustrates the learning curves of accuracy and loss for GAMED and its four modules over 11 epochs of training on the Fakeddit and Yang datasets. The learning curves on the Fakeddit dataset show steady improvement, with accuracy gradually increasing and loss sharply decreasing before stabilizing, indicating efficient learning and convergence. In contrast, the Yang dataset exhibits more variability, likely due to its smaller size and higher diversity and complexity of samples, which makes it more challenging for the model to consistently capture and learn patterns. The accuracy curves fluctuate before stabilizing, and although the loss initially decreases, it also shows greater volatility, reflecting the model’s difficulty in consistently optimizing when confronted with diverse characteristics of the dataset. Despite these fluctuations, both datasets demonstrate that the models are learning and improving. Notably, although there are some fluctuations during training, the fusion module (MM) shows a generally stable trend in both accuracy and loss across these datasets, with no substantial improvements. This validates our view that the progress of the overall architecture is primarily driven by the synergy of multiple modalities, particularly the unimodal distinctiveness and discriminative power, while early fusion contributes only minimally.

\begin{table}
\scalebox{0.804}
{
\begin{tabular}{lc}
\hline
\textbf{Test} & \textbf{Accuracy} \\ \hline
MM & 0.614 \\
IP & 0.820 \\
IS & 0.855 \\
IS+IP & 0.885 \\
T  & 0.901 \\
IP+T & 0.906 \\
IS+T & 0.909 \\
IS+IP+T & 0.914 \\ \hline
\end{tabular}
}
\quad
\scalebox{0.804}
{
\begin{tabular}{lc}
\hline
\textbf{Test} & \textbf{Accuracy} \\ \hline
MM-only ComputeStats & 0.884 \\
w/o AdaIN & 0.908 \\
w/o Coarse Prediction & 0.916 \\
Replace IRNv2 with Inception-v3 & 0.924 \\
Replace MMoE-Pro with ViT block & 0.926 \\
w/o Consistency Learning & 0.926 \\
Replace ERNIE2.0 with BERT & 0.927 \\
Replace MMoE-Pro with MMoE & 0.928 \\
Replace MAE-ViT with ViT & 0.931 \\
w/o Veto Voting & 0.933 \\
Replace ERNIE2.0 with ERNIE1.0 & 0.936 \\ \hline
GAMED & \textbf{0.939} \\ \hline
\end{tabular}
}
\caption{Ablation results of key modules and components in the GAMED design, tested on Fakeddit.}
\label{ablation-results}
\vspace{-6mm}
\end{table}

\subsection{Ablation Study}
\noindent \textbf{Removing Individual Modules:} In Table \ref{ablation-results}, we present the ablation results. By removing modules, we find that the text module performs the best among all individual modules, achieving an accuracy of 90.1\%, which even exceeds many excellent fake news detection models. Next are IS and IP, but their combined performance only reached 88.5\%, still trailing the text module by 1.6\%. The worst-performing individual module is MM, with a peak accuracy of only 61.4\%, demonstrating the effectiveness of our weakened fusion module design. However, the combination of individual IS and T using the same data achieved an accuracy of 90.9\%, significantly outperforming the MM module by 29.5\%. More importantly, GAMED maintained strong detection capabilities despite the poor performance of the MM module. This further supports our view that the contributions of unimodal distinctiveness and their cross-modal synergies to model performance outweigh standalone modal fusion. Finally, compared with the high accuracy of GAMED, these removals prove that no single modality or any combination can reach the overall performance of GAMED.

\noindent \textbf{Knowledge Enhancement:} We used BERT to replace ERNIE to process the text data, but the result dropped by 1.2\%. Although both two models are based on similar transformer architecture, the advantage of ERNIE is that it further incorporates a structured knowledge graph to enhance the understanding of facts and relations. In addition, we used ERNIE1.0 instead of ERNIE2.0, and the result dropped by 0.3\%. This is because ERNIE2.0 introduced a more complex knowledge increment strategy and larger knowledge parameters than ERNIE1.0 during pre-training. In the previous individual module removal, we found that the performance of a single text module was stronger than the combined performance of images modules, and the initial performance gap between IP and IS was large; but after the enhancement of the text module, not only the performance was greatly improved, both exceeded 90\%; however, the gap between IP and IS became very small. ERNIE's victory highlights that external knowledge integration is crucial to enhancing the overall performance in complex tasks, i.e., multimodal fake news detection. In addition, to measure the impact of different feature extractors, we used InceptionV3 instead of Inception-ResNet-v2 and ViT instead of MAE-ViT, which resulted in a 1.5\% and 0.8\% drop in accuracy for GAMED, respectively.

\noindent \textbf{Expert Network:} We replaced MMoE-Pro with standard MMoE, resulting in a 1.1\% drop in GAMED's accuracy. This decline is due to MMoE-Pro's enhancements in feature sharing and task relationship modelling, which allow the model to effectively process and fuse multimodal data. We replaced the MMoE-Pro networks with the ViT blocks for feature refinement, but the accuracy dropped by 1.3\%. This suggests that sharing information is beneficial for enhancing representations in our task, while the ViT blocks reduced the model's ability to flexibly and dynamically select the most relevant features for the task. We removed the coarse prediction step and the accuracy of GAMED dropped by 2.3\%, which proves that the constraint of expert opinion is useful for evaluating the importance of modalities. 

\noindent \textbf{Adaptive Adjustment:} We experimented with the mean and standard deviation of MM as the input of AdaIN to adjust the distribution of other unimodal modules. We found that the accuracy of GAMED dropped by 5.5\%, which was a significant drop. In contrast, when we removed the consistency learning, the accuracy of GAMED dropped by 1.3\%. This suggests that although consistency learning can improve the model's ability to identify fake news, prioritizing it does not bring greater benefits. Instead, it undermines the contribution of unimodal discriminability to the overall performance. We then removed the AdaIN setting for all modalities, and the accuracy of GAMED dropped by 3.1\%. This is because AdaIN can adaptively adjust the feature distribution of different modalities to ensure that the most valuable features are provided for the interaction link.

\noindent \textbf{Decision Making:} We used concatenation-based late fusion instead of veto voting, resulting in a 0.6\% drop in the accuracy of GAMED. This is because our carefully designed cross-modal interaction rules for veto voting can dynamically adjust the key modalities to ensure that the most reliable predictions have the greatest impact on the final decision. At the same time, other methods lose such flexibility and are more susceptible to noise.

\begin{figure}
    \centering
    \includegraphics[scale=0.306]{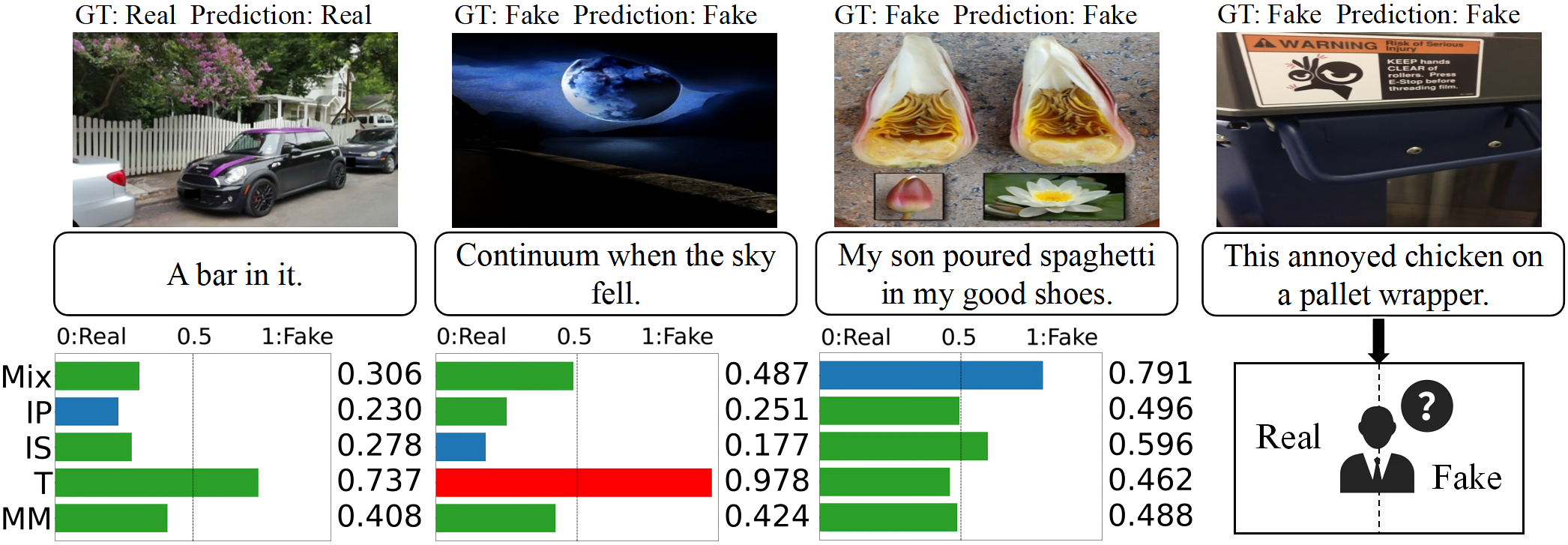} %
    \caption {The interpretability of GAMED is illustrated through the decision-making process. The images and text are sourced from real examples in the Fakeddit test set.}
    \label{fig:Interpretability}
    \vspace{-2mm}
\end{figure}

\subsection{Qualitative Analysis}
As shown in Figure~\ref{fig:Interpretability}, we demonstrate the interpretability of the decisions made by GAMED using real examples. We randomly selected three prediction results on the Fakeddit test set and traced the corresponding samples. From left to right, following the veto voting rule in Section \ref{our-model}, the prediction result of the first sample is ``Real'', this is because the confidence of several modalities is between the pre-set low threshold \( \theta_{\text{low}} \) and the high threshold \( \theta_{\text{high}} \), so the initial concatenated prediction \(P_{\mathit{mix}}\) is directly used as the final prediction. \(P_{\mathit{mix}}\) represents the result after the multimodal features are concatenated, and it makes full use of the complementarity of each modality to make the prediction result more comprehensive and reliable. The second sample shows that since the output confidence of the text module \(P_t\) is higher than the high threshold \( \theta_{\text{high}} \) and higher than \(P_{\mathit{mix}}\), the prediction of the initial \(P_{\mathit{mix}}\) is replaced with the prediction \(P_t\) as the final decision. The text modality provides extremely reliable information in this context. We can maximize the use of this reliable information and improve the accuracy of the final decision. The third sample reflects the decision made by GAMED when the output confidence of the image pattern module \(P_{\mathit{ip}}\) is lower than the low threshold \( \theta_{\text{low}} \). This is because too low confidence means that the information of this modality may be unreliable or misleading, even if it belongs to the majority class. By ignoring this unreliable prediction and comprehensively reconsidering the combination of the highest output in all modalities and the concatenated output, the robustness of the final decision is ensured. The fourth sample simulates the black-box decision-making process of many current models, in which the model cannot clearly explain the specific reasons for its decision. This may not only lead to a decrease in user trust in the model's prediction results but also make it difficult to debug and improve effectively when errors occur. In addition, GAMED's modal-decoupling design is also used for interpretability. For example, when AdaIN adaptively adjusts the feature distribution of different modules, we can judge the contribution of each modality and its discriminative features to the prediction. In contrast, those black-box models do not have a clear explanation path when processing input data. This means that the internal working mechanism of the model is invisible to users and developers, resulting in people being unable to understand or verify the reasoning process behind it even if the model makes a correct classification.

\section{Conclusions}
This paper developed GAMED -- a novel architecture that significantly improves fake news detection. GAMED overcomes the shortcomings of current multimodal approaches through a dynamic mechanism of modal decoupling and cross-modal synergy. It embeds the benefits of semantic information encoded in knowledge graphs into the whole workflow from pre-trained language models. Feature selection is performed jointly by a mixture of experts, accompanied by subsequent adaptive distribution adjustment, progressively refining the feature representation of each modality in the pipeline and enhancing its discriminability and distinctiveness. Finally, a flexible and transparent decision process is introduced. Our experiments on benchmark datasets, Fakeddit and Yang, show that GAMED improves upon recent top-performing models in detection accuracy. Future work would explore adding more modalities such as audio or video for a more holistic analysis of fake news.

\section{Acknowledgment}
This work was supported by the Alan Turing Institute/DSO grant on improving multimodality misinformation detection through affective analysis. We gratefully acknowledge NVIDIA for providing computational resources through its NVIDIA Academic Hardware Grant Program 2021. Additional support was provided by the Interdisciplinary Research Pump-Priming Fund, University of Southampton.

\newpage

\section{Ethical Considerations}
Some of the key ethical considerations include addressing the issues when models can perpetuate existing societal biases if the training data is biased. This can lead to discriminatory outcomes, such as unfairly targeting certain groups or individuals. Besides that, different cultures have varying norms and understandings of truth and fake news. Models trained on data from one culture may not perform well or ethically in another. Another fundamental challenge lies in overly aggressive detection models that could lead to the suppression of legitimate speech, particularly for marginalized voices or those critical of authority. Incorrectly flagging accurate information as fake news can damage reputations and stifle public discourse. Our goal in this work is to develop a model that could understand how it reached its conclusions. Black-box models make it difficult to identify and address biases. Overreliance on automated detection systems could erode trust in traditional media and journalism.

\balance

\end{document}